\PassOptionsToPackage{final}{graphicx}
\documentclass{bmvc2k}

\usepackage{booktabs}
\usepackage{multirow}
\usepackage{graphicx}
\usepackage{amssymb}
\usepackage{pifont}
\usepackage{makecell}
\usepackage[table]{xcolor}
\usepackage{wrapfig}
\usepackage{float}

\newcommand{\bcirc}[1]{\ding{\numexpr171+#1\relax}}

\title{\textit{Not All Patches Are Equally Forgettable}: Spatially Localized Domain Unlearning in Vision-Language Models}

\addauthor{Akanksha Singh}{akanksha23@iiserb.ac.in}{1}
\addauthor{Vinod K. Kurmi}{vinodkk@iiserb.ac.in}{1}

\addinstitution{
Indian Institute of Science Education and Research Bhopal, India
}

\runninghead{A. SINGH, V.K. KURMI}{NOT ALL PATCHES ARE EQUALLY FORGETTABLE}

\begin{document}
\maketitle
\begin{abstract}
\noindent Pre-trained vision-language models (VLMs) exhibit strong cross-domain recognition performance even without additional training. However, this robustness can also preserve undesirable domain-specific behavior, as domain-related and semantic information often remain entangled within the learned representation space, making selective domain unlearning challenging. Existing approaches typically address this problem through latent-space
disentanglement and prompt- or feature-level interventions, without directly attributing and attenuating individual patch-token contributions. However, here we suggest that rather than uniformly suppressing the full representation, it may be more effective to exploit the spatial structure of vision transformers to localize and suppress patch regions that contribute disproportionately to forget-domain prediction. Patches that strongly influence forget-domain prediction may not be equally important for semantic recognition, suggesting that forgetting should be guided according to the domain contribution of different visual regions. Specifically, we propose a two-stage patch-selective framework that first estimates patch-level domain sensitivity and then selectively attenuates patches whose contribution to forget-domain prediction is stronger than their semantic utility. We evaluate our framework on Office-Home, Mini DomainNet, and DomainNet. Experimental results demonstrate improved forgetting-retention tradeoffs compared to prior methods while improving retained-domain recognition by up to 3.8\%. Additional evaluations under visually overlapping and unseen-domain settings further demonstrate improved robustness under distribution shift.
\end{abstract}
\vspace{-0.6em}
\section{Introduction}
\label{sec:intro}
\vspace{-0.7em}
Pre-trained vision-language models (VLMs) have demonstrated strong cross-domain generalization performance, enabling robust zero-shot and few-shot recognition across diverse visual domains \cite{radford2021learning,jia2021scaling,li2022blip,yu2022coca}. Large-scale image-text pretraining allows these models to learn semantically transferable representations that remain effective under substantial distribution shifts. However, this same broad transferability may also preserve undesirable domain-specific behavior, where the model relies on domain-dependent cues such as artistic style, texture statistics, background context, or rendering artifacts while making predictions \cite{varma2024ravl,zhang2024amend,chi2024adapting}. For example, a VLM trained on large-scale web data may associate repetitive stylistic or domain-specific visual patterns with particular object categories, affecting predictions to partially depend on domain cues rather than purely object information \cite{varma2024ravl,zhang2024amend}.

Recent works \cite{kawamura2026approximate} formulate this setting as \textit{approximate domain unlearning}, where the objective is to suppress recognition behavior associated with a target domain while preserving recognition on the remaining domains. Existing approaches disentangle domain-specific information in the latent space, with ADU~\cite{kawamura2026approximate} using intermediate patch features to condition image-level prompt generation. This formulation is well motivated by the observation that domain distributions become highly entangled within a shared feature space. Meanwhile, vision transformers, which are widely adopted as visual backbones \cite{dosovitskiy2020image}, naturally represent images as spatially localized patch tokens prior to global aggregation, providing an inherent spatial representation structure within the model itself.

\begin{wrapfigure}{r}{0.50\linewidth}
\vspace{-1.0em}
\centering
\includegraphics[width=\linewidth]{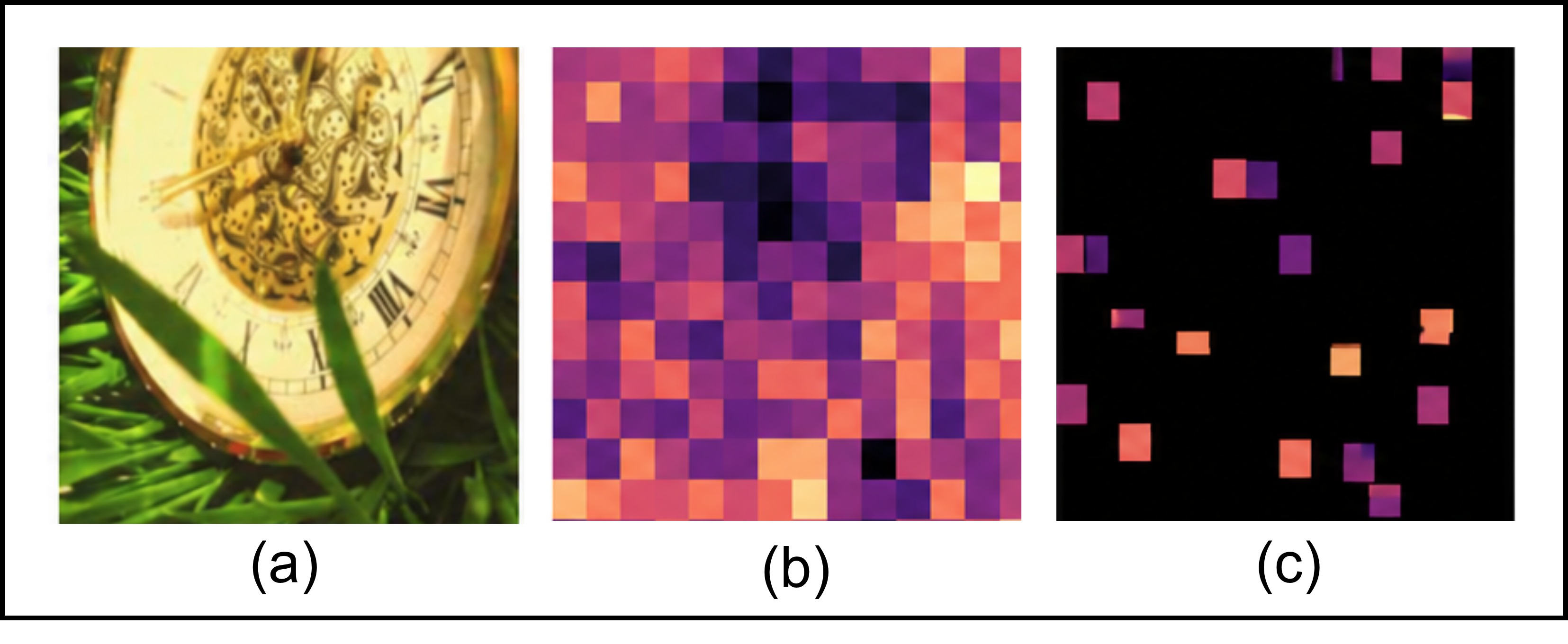}
\vspace{-2em}
\caption{\textbf{Patch-level visualization of forget-domain contribution.} 
(a) Input image, (b) patch-level domain sensitivity heatmap where 
yellow: high domain contribution and purple: low 
contribution, and (c) top domain-predictive patches. \textit{(Best viewed in color.)}}
\label{fig:intro_patch}
\vspace{-1.2em}
\end{wrapfigure}
\vspace{0.3em}

\vspace{0.2em}
\noindent Prior studies show that different patches capture different forms of visual evidence, including object structure, texture, style, and background statistics \cite{raghu2021vision,naseer2021intriguing,chefer2021transformer}. As a result, domain-related information may not be distributed uniformly across patch tokens, where some regions contribute more strongly to domain recognition while others primarily support semantic recognition shared across domains, as illustrated in Fig.~\ref{fig:intro_patch}.  This observation becomes particularly important for approximate domain unlearning, where our objective is to suppress recognition behavior associated with a target domain while preserving semantic recognition on other domains. This naturally motivates exploiting the spatial structure of vision transformers to guide forgetting according to how strongly individual patches contribute to domain predictions.

Importantly, domain-sensitive patches are not necessarily independent of semantic information. A patch may simultaneously encode both domain-specific and semantically useful information while contributing disproportionately to domain prediction and semantic recognition (in Section~\ref{sec:motive}). For example, texture, rendering style, or background context may strongly influence domain prediction while contributing less to semantic object recognition, whereas object-centric regions may show the opposite behavior. These observations suggest that forgetting should instead be guided according to the relative contribution of each patch to domain discrimination and semantic recognition. Building upon this, in this work, we propose a two-stage localized patch-guided framework that localizes domain effects within intermediate patch representations based on their relative contribution to domain prediction and semantic recognition.

\vspace{0.2em}
\noindent\bcirc{1}\ \textbf{Stage I: Patch-level sensitivity estimation. } We train a lightweight patch-level domain scorer over intermediate ViT tokens to estimate how strongly each patch contributes to forget-domain prediction. This produces spatial importance estimates that identify domain-dominant visual regions prior to global aggregation.

\vspace{0.2em}
\noindent\bcirc{2}\ \textbf{Stage II: Localized Patch Attenuation. }  We estimate the semantic utility of each patch using its alignment with class-discriminative text representations and combine it with the corresponding domain sensitivity score. Patches whose domain contribution disproportionately exceeds their semantic contribution are selectively attenuated before forwarding the representation through the remaining transformer layers. Extensive experiments on Office-Home \cite{venkateswara2017deep}, Mini DomainNet \cite{peng2019moment}, and DomainNet \cite{peng2019moment} demonstrate that our localized patch-guided framework consistently achieves stronger forgetting- retention tradeoffs. We additionally evaluate robustness under visually overlapping domain settings on Mini DomainNet and unseen domain generalization settings on DomainNet, where our method tends to preserve generalization capabilities. 
\textbf{Our key contributions can be summarized as follows:}
    \vspace{-0.7em}
\begin{itemize}
    \item We formulate approximate domain unlearning through explicit token-level attribution and intervention, estimating localized forget-domain sensitivity and directly modifying selected patch representations before global aggregation.

    \vspace{-0.7em}
    \item We propose a localized patch-guided domain suppression framework that selectively
    attenuates patches that contribute disproportionately to domain prediction while preserving
    semantically informative regions.

     \vspace{-0.7em}
    \item We introduce a patch-level scoring mechanism that jointly estimates domain sensitivity and semantic utility to identify patches whose contribution to forget-domain prediction disproportionately exceeds their semantic contribution.
    
      \vspace{-0.7em}
    \item Extensive experiments across multiple multi-domain benchmarks demonstrate that localized patch-level intervention achieves stronger forgetting-retention tradeoffs than prior global suppression approaches in CLIP-based vision-language models.
\end{itemize}

\vspace{-1.8em}
\section{Related Work}
\label{sec:rel}
\vspace{-0.5em}
\noindent \textbf{Machine Unlearning in Vision-Language Models.} Machine unlearning studies how to remove selected data, classes, or attributes from a trained model without full retraining \cite{cao2015towards, bourtoule2021machine, sekhari2021remember}. As retraining-based deletion is computationally expensive, most methods pursue approximate unlearning through gradient ascent, parameter suppression, saliency estimation, distillation, or representation editing \cite{golatkar2020eternal, graves2021amnesiac, chundawat2023can, foster2024fast}. In vision models, class-level and sample-level forgetting are commonly achieved by increasing forget-set loss while maintaining retain-set accuracy or by suppressing parameters \cite{golatkar2020forgetting, warnecke2021machine}. Extending unlearning to vision-language models is substantially more challenging because image-text pretraining produces broad semantic generalization and entangled multimodal representations \cite{radford2021learning, jia2021scaling, zhou2022learning}. Existing VLM unlearning approaches therefore rely on prompt tuning, latent representation editing, activation steering, sparse concept suppression, or parameter-level adaptation to suppress target behavior \cite{kawamura2026approximate, gandikota2023erasing, geng2025sauce}. ADU~\cite{kawamura2026approximate} additionally uses intermediate patch representations to condition instance-wise prompt generation.

\vspace{0.4em}
\noindent \textbf{Domain Adaptation and Domain Generalization. }  A large body of recent work in domain adaptation and domain generalization studies how domain-specific information is encoded and manipulated within visual and vision-language representations under distribution shift \cite{zhou2022domain, zhang2024vision}. Existing approaches reduce domain discrepancy through adversarial alignment, prompt tuning, source-free adaptation, adapters, or test-time optimization \cite{long2018conditional,zhou2022conditional,khattak2023maple,derakhshani2023bayesian}. These studies collectively show that pretrained VLM representations retain substantial domain-specific structure even under strong multimodal pretraining. However, their primary objective is to improve transferability by globally suppressing domain discrepancy across domains. In contrast, approximate domain unlearning \cite{kawamura2026approximate} requires selectively weakening recognition behavior associated with a specified target domain while preserving semantic recognition on retained domains. This shifts the focus from learning globally domain-invariant representations to identifying which components of the representation contribute disproportionately to forget-domain prediction.

\vspace{0.4em}
\noindent \textbf{Spatial Representation Analysis in Vision Transformers. } Vision transformers represent images as sequences of spatial patch tokens before global aggregation \cite{dosovitskiy2020image,touvron2021training}. Prior studies show that different patch tokens encode heterogeneous visual evidence, including object structure, texture, style, background statistics, and shortcut cues \cite{raghu2021vision,naseer2021intriguing,chefer2021transformer}. These observations have motivated token pruning, attention localization, and patch-level interpretability methods that selectively suppress uninformative regions while preserving task-relevant structure \cite{rao2021dynamicvit,yin2022vit}. Related works further suggest that visual evidence is often distributed unevenly across spatial regions rather than uniformly throughout the image \cite{abnar2020quantifying}.\\
\indent These findings motivate our \textit{\textbf{central hypothesis}} that forget-domain evidence in VLMs may likewise be spatially localized within intermediate patch representations, enabling
domain unlearning through direct intervention on localized domain-discriminative regions
rather than uniformly editing the representation space.

\vspace{-1.0em}
\section{Motivation} 
\label{sec:motive}
\vspace{-0.7em}
We empirically validate our central hypothesis that some patch tokens contribute disproportionately to forget-domain prediction. We train a lightweight linear scorer over intermediate ViT tokens with image-level domain supervision \textit{ (architectural and optimization details are provided in Section~\ref{sec:proposed_method})}. Using the scorer, we study how strongly intermediate patches contribute to domain prediction and semantic recognition. We formulate research questions as:
\vspace{0.7em}
\begin{wraptable}{r}{0.56\linewidth}
\vspace{-0.8em}
\centering
\scriptsize
\caption{\textbf{Effect of selective patch suppression on Office-Home \textit{(|$D_f|=1$)}.} Top-K suppresses the highest domain-scoring patches ($\textit{K}$ expressed as \% of total tokens).}
\vspace{0.2em}
\setlength{\tabcolsep}{10pt}
\begin{tabular}{lcc}
\toprule
\textbf{Setting} & \textbf{Forget} & \textbf{Retain} \\
& \textbf{Acc. $\downarrow$} & \textbf{Acc. $\uparrow$} \\
\midrule
All patches & 63.23 & 78.40 \\
\rowcolor{red!8}Top-$K$ patches & 37.20 & 77.29 \\
\bottomrule
\end{tabular}
\label{tab:patch_domain_analysis}
\vspace{-1.4em}
\end{wraptable}
\noindent\textbf{RQ-1: Do some patches disproportionately contribute to forget-domain prediction?}
We evaluate whether selectively suppressing only a small subset of highly domain-sensitive patches achieves stronger forgetting than uniformly suppressing all patches. As shown in Table~\ref{tab:patch_domain_analysis}, Top-K suppression ($K{=}20$, approximately $10\%$ of 196 tokens) substantially reduces forget-domain accuracy (63.23 $\rightarrow$ 37.20) while incurring only a negligible drop in retained-domain accuracy (78.40 $\rightarrow$ 77.29). These results support the hypothesis that a small subset of highly domain-sensitive patches
contributes disproportionately to forget-domain prediction.

\vspace{0.5em}
\begin{wrapfigure}{r}{0.5\linewidth}
\vspace{-1.0em}
\centering
\includegraphics[width=\linewidth]{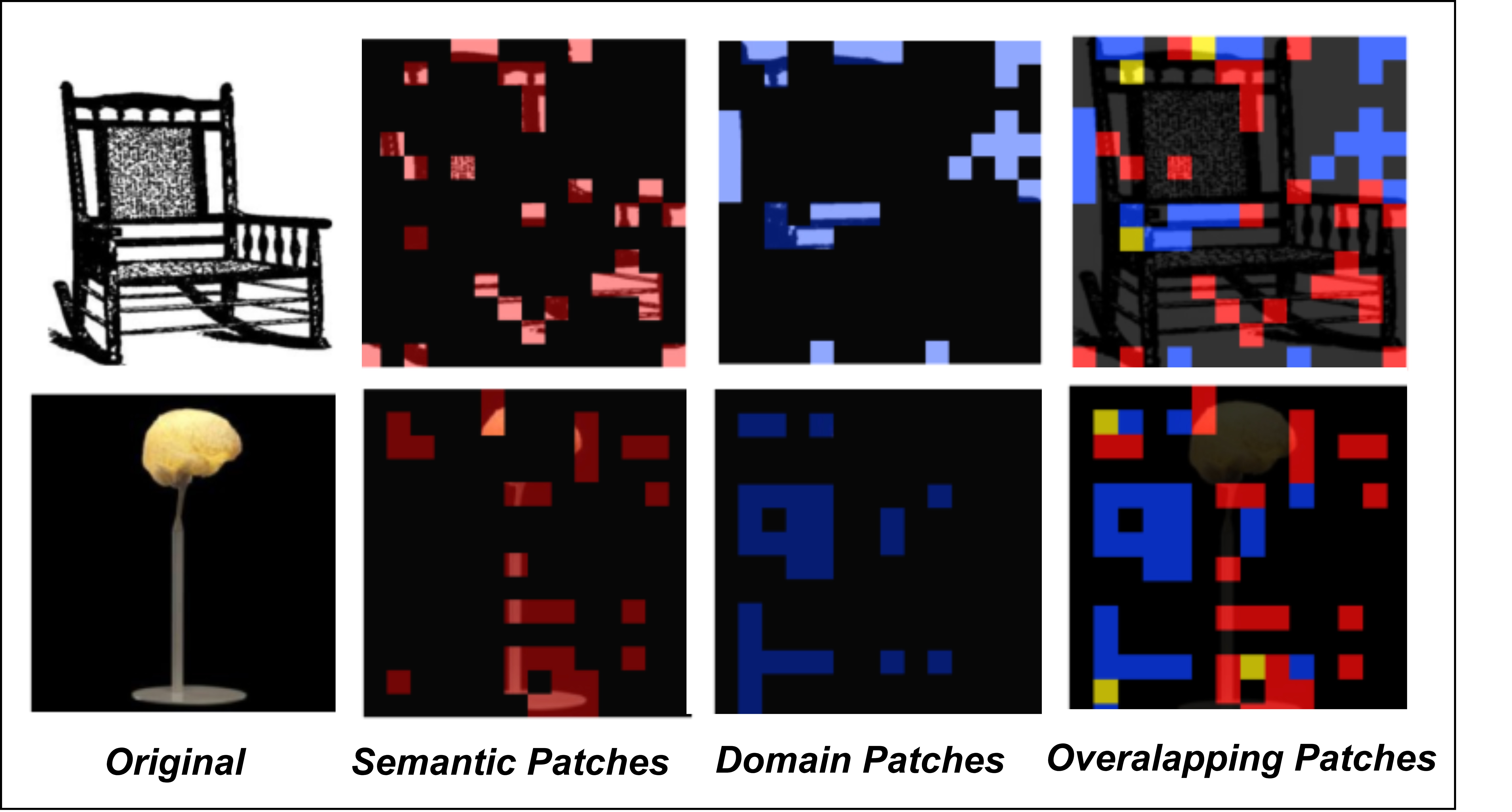}
\vspace{-1.8em}
\caption{\textbf{Patch-level domain contribution visualization.} (a) input image, (b) red: semantic patches, (c) blue: domain patches, (d) yellow: overlapping patches (\textit{best view in color).}}
\label{fig:motivation_patch}
\vspace{-1.2em}
\end{wrapfigure}

\noindent\textbf{RQ-2: Do domain-predictive patches spatially overlap with class-informative patches?}
Fig.~\ref{fig:motivation_patch} shows that class-informative patches are primarily concentrated around object-centric regions, whereas highly domain-predictive patches only partially overlap with them and additionally appear in surrounding contextual regions. This suggests only partial alignment between domain and semantic evidence across patches. Dataset-level overlap analysis further supports this observation (\textit{Supplementary}). The non-overlapping domain-predictive regions motivate selective suppression guided by relative domain contribution rather than uniform perturbation across all patches. Building on these observations, we propose a localized patch-guided framework for approximate domain unlearning. Rather than uniformly perturbing the global representation, our method selectively attenuates patches according to their relative contribution to forget-domain prediction and semantic recognition.
\begin{figure*}[t]
    \centering
    \includegraphics[width=\textwidth]{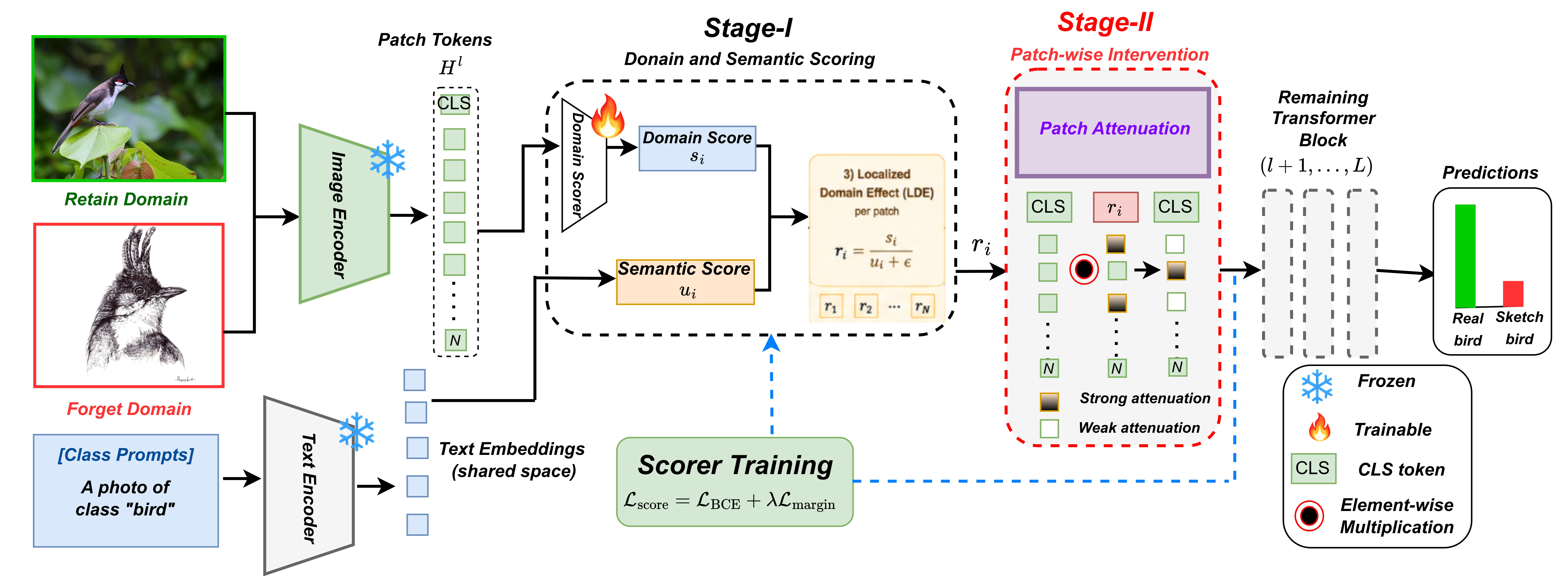}
    \vspace{-1em}
    \caption{\textbf{Overview of the proposed localized patch-guided domain unlearning framework.}
    At an intermediate layer $\ell$, patch representations are evaluated using a trainable domain scorer and frozen semantic utility from class-text embeddings. Their domain and semantic scores, $s_i$ and $u_i$, define the localized domain effect $r_i$, which determines patch-wise attenuation. The modified tokens are then propagated through the remaining frozen transformer blocks for prediction. Only $g_\phi$ is trained using $\mathcal{L}_{\mathrm{score}}=\mathcal{L}_{\mathrm{BCE}}+\lambda\mathcal{L}_{\mathrm{margin}}$.}
    \label{fig:pipeline}
    \vspace{-1.4em}
\end{figure*}

\vspace{-1.2em}
\section{Proposed Method}
\label{sec:proposed_method}
\vspace{-0.6em}
We consider approximate domain unlearning in Vision-Language Models (VLMs). Let $\mathcal{T}=\{(x,y,d)\}$ denote the training set, where \(x\) is an input image, \(y\) is the semantic class label, and \(d\) is the domain label. Given retained domains \(D_r\) and forget domains \(D_f\), our goal is to modify a pretrained VLM \(f_{\theta}\) such that classification performance is preserved on samples from \(D_r\) while being reduced on samples from \(D_f\). Given an image \(x\), the vision transformer represents it as a sequence of patch tokens,
\vspace{-0.4em}
\[
\small
H^l(x)=\{h_1^l,\dots,h_N^l\},
\vspace{-0.6em}
\]
where \(h_i^l\) denotes the representation of the \(i\)-th patch token at layer \(l\). The final prediction is obtained after passing
intermediate patch representations through the subsequent transformer layers, token aggregation, and classification head given as
\vspace{-0.4em}
\[
\small
p_{\theta}(\hat{y}\mid x)=g_{\theta}\!\left(H_l(x)\right)
\vspace{-0.6em}
\]
where $g_{\theta}(\cdot)$ denotes the classifier head. Since different patches may contribute disproportionately to forget-domain prediction and semantic recognition, uniformly perturbing all tokens may unnecessarily suppress semantically useful information. We therefore formulate approximate domain unlearning as a selective patch-level suppression problem. Given an intermediate patch representation $h_i^l \in H^l(x)$, our objective is to construct modified patch representations
\vspace{-0.6em}
\[
\small
\widetilde{H}^l(x)=\{\widetilde{h}_1^l,\dots,\widetilde{h}_N^l\},
\vspace{-0.4em}
\]
where patches are selectively attenuated according to their relative forget-domain contribution prior to global aggregation. The resulting representation aims to preserve recognition performance on samples from \(D_r\) while reducing recognition performance on \(D_f\) samples.

\vspace{0.7em}
\noindent \underline{\textbf{\textit{Method Overview}}}  Building upon the formulation above, we propose a two-stage localized patch-guided framework for approximate domain unlearning shown in Fig.~3.
\textcircled{1} \textbf{Stage I:} We first estimate the relative contribution
of each patch to forget-domain prediction and semantic recognition and term it
the Localized Domain Effect (LDE). \textcircled{2} \textbf{Stage II:} The estimated LDE scores are then used to
selectively attenuate patches with higher relative forget-domain contribution
before forwarding the representations through the remaining transformer layers.

\vspace{-0.5em}
\subsection{Stage I: Patch-level sensitivity estimation.}
\vspace{-0.5em}
Our goal in Stage I is to estimate the relative contribution of each intermediate patch token towards forget-domain prediction and semantic recognition individually. To achieve this, we first compute a patch-level domain sensitivity score and subsequently estimate the semantic utility of each patch using the pretrained image-text alignment space.

\vspace{0.5em}
\noindent \underline{\textit{\textbf{(1) Patch-Level Domain Sensitivity Estimation.}}}
To estimate how strongly each patch helps in predicting the forget domain, we introduce a lightweight patch scoring function $g_\phi:\mathbb{R}^{d}\rightarrow\mathbb{R}.$ For a given patch representation $h^{l}_{i}$, the scorer produces a scalar domain sensitivity score
\vspace{-0.4em}
\[
\small
s_i = \sigma(g_\phi(h_i^\ell)),
\
\]
where larger values indicate stronger domain sensitivity. Since only image-level domain supervision is available, patches from forget-domain images $D_f$ inherit positive labels, while patches from retained-domain images $D_r$ inherit negative labels. The scorer is optimized over $N$ patch representations using binary cross-entropy,
\vspace{-0.6em}
\[
\small
L_{\mathrm{BCE}}
=
-\frac{1}{N}
\sum_{i=1}^{N}
\left[
z_i\log s_i
+
(1-z_i)\log(1-s_i)
\right],
\vspace{-0.6em}
\]
where $z_i\in\{0,1\}$ denotes the inherited patch-level domain label. Although supervision is taken from image-level labels, scoring is performed over contextualized patch representations. This is similar to weakly supervised patch-level estimation strategies explored in prior transformer-based localization and adaptation works~\cite{zhu2023patch, zhu2024boosting}. Consequently, patches that consistently encode forget-domain cues get larger sensitivity scores, whereas patches weakly associated with the forget domain or shared across retained and forget domains receive comparatively lower scores. To further increase the relative separation between forget-domain and retained-domain patches, we additionally introduce a margin-based ranking objective
\vspace{-0.8em}
\[
\small
L_{\mathrm{margin}}
=
\sum_{(i,j)\in P_f\times P_r}
\max\left(0,m-(s_i-s_j)\right),
\vspace{-0.4em}
\]
where $P_f$ and $P_r$ denote randomly sampled forget-domain and retained-domain patch sets, respectively, and $m$ is a margin hyperparameter. This objective encourages forget-domain patches to obtain larger sensitivity scores than retain domain patches. The final domain sensitivity objective becomes
\vspace{-0.4em}
\[
\small
L_{\mathrm{score}}
=
L_{\mathrm{BCE}}
+
\lambda L_{\mathrm{margin}}.
\vspace{-0.4EM}
\]

\vspace{0.6em}
\noindent \underline{\textit{\textbf{(2) Semantic Utility Estimation.}}}
While the domain sensitivity score captures forget-domain contributions, certain high-scoring patches may still encode useful semantic structure. We therefore estimate semantic utility using the pretrained image-text alignment space \cite{mukhoti2023open}. Each patch representation is projected into the shared embedding space as $\hat{h}_i = P h^{l}_{i}$, where $P$ denotes the frozen visual projection matrix. Let $\{t_1,\ldots,t_C\}$ denote normalized text embeddings corresponding to semantic class descriptions, where $c \in \{1,\ldots, C\}$ indexes the class embeddings. The semantic utility of a patch $h^{l}_{i}$ is then computed as
\vspace{-0.4em}
\[
\small
u_i
=
\max_c
\cos(\hat{h}_i,t_c).
\vspace{-0.6em}
\]
Using the maximum similarity across classes preserves semantically meaningful patches irrespective of class identity. As individual patches may encode semantic structures shared across multiple classes, using maximum similarity across class embeddings therefore preserves semantically meaningful regions instead of restricting semantic preservation to only the ground-truth category.

\vspace{0.6em}
\noindent \underline{\textit{\textbf{(3) Localized Domain Effect.}}} The estimated domain sensitivity and semantic utility are subsequently combined into a unified Localized Domain Effect (LDE) score 
\vspace{-0.4em}
\[
r_i
=
\frac{s_i}{u_i+\epsilon},
\vspace{-0.6em}
\]
where $\varepsilon > 0$ ensures numerical stability. This formulation follows from the requirement that the score should increase with domain sensitivity while being proportionally suppressed as semantic utility increases. Consequently, semantic utility naturally acts as a normalizing denominator rather than an independent additive term. Thus, patches dominated by forget-domain sensitivity receive larger LDE scores, whereas semantically informative regions receive comparatively smaller scores. Here, ``forgettable'' refers to patches with high relative LDE rather than an intrinsic property of the patches.

\vspace{-1em}
\subsection{Stage II: Localized Patch Attenuation}
\vspace{-0.5em}
Stage II suppresses intermediate patch representations according to their estimated Localized Domain Effect (LDE) scores. Given patch-level scores $\{r_1,\ldots,r_N\}$, the objective is to selectively weaken patches whose contribution to forget-domain prediction is disproportionately higher than their semantic contribution. As the patch-specific suppression behavior is already encoded through the LDE scores, suppression is applied directly over intermediate patch representations without introducing additional learnable suppression modules. For each intermediate patch representation $h_i^l$, attenuation is applied as
\vspace{-0.7em}
\[
\small
\tilde{h}_i^l=
\begin{cases}
(1-\alpha \bar{r}_i)h_i^l, & r_i>\tau,\\
h_i^l, & r_i\leq\tau,
\end{cases}
\vspace{-0.6em}
\]
where $\alpha>0$ controls the overall suppression strength, $\tau$ denotes the LDE threshold, and $\bar{r}_i$ denotes the normalized LDE score used for stable attenuation. The transformed patch sequence becomes
\vspace{-0.4em}
\[
\small
\tilde{H}^l(x)=\{\tilde{h}_1^l,\ldots,\tilde{h}_N^l\}.
\vspace{-0.4em}
\]
Patches with low LDE scores are preserved, while patches with high LDE scores are selectively attenuated before forwarding the representations through the remaining transformer layers. Since suppression is applied prior to global aggregation, the intervention weakens localized forget-domain contributions without uniformly perturbing the global representations.

\vspace{-1.2em}
\section{Results and Analysis}
\vspace{-0.6em}
\subsection{Experimental Setup}
\label{sec:exp}
\vspace{-0.35em}
\noindent\textbf{Datasets. } We evaluate on four multi-domain benchmarks commonly used in domain adaptation and domain generalization—ImageNet and ImageNet-Sketch \cite{deng2009imagenet,wang2019learning}, Office-Home \cite{venkateswara2017deep}, Mini DomainNet \cite{zhou2021domain}, and DomainNet \cite{peng2019moment}. Following \cite{kawamura2026approximate}, ImageNet-1K and ImageNet-Sketch are treated as two distinct domains over 1000 shared classes. Office-Home contains 65 classes across Art, Clipart, Product, and Real-World domains, while Mini DomainNet contains 126 classes across Clipart, Painting, Real, and Sketch. Additional DomainNet results are provided in the \textit{Supplementary}. Unless otherwise stated,
we use eight labeled samples per domain; supervision-budget results are in the Supplementary.

\vspace{0.35em}
\noindent\textbf{Implementation Details and Evaluation Metrics. } We use frozen CLIP ViT-B/16 \cite{dosovitskiy2020image} with the text prompt ``a photo of a [class]''. Intermediate patch tokens from transformer layer \textit{5} are used for patch-level scoring and suppression. Only the lightweight patch scorer is optimized using SGD for 50 epochs with a learning rate of 0.0025. Following \cite{kawamura2026approximate}, we report \textbf{Mem}, retained-domain accuracy; \textbf{\textit{For}}, forget-domain error; and \textbf{\textit{H}}, their harmonic mean. Higher values indicate better unlearning performance. Results are averaged over three random seeds. Unless otherwise stated, we use $\alpha=1.0$, $\lambda=0.5$, $\tau=0.3$ across all experiments.

\vspace{0.35em}
\noindent\textbf{Baselines. } Existing approximate domain unlearning methods for VLMs do not explicitly perform selective forgetting at the patch representation level. We therefore primarily compare against ADU \cite{kawamura2026approximate}, which uses intermediate patch features
to condition instance-wise prompts for domain suppression in the same setting. We additionally compare against recent CLIP adaptation methods, including LP++ \cite{huang2024lp++} and CLIPFit \cite{li2024vision}, as well as the VLM unlearning method Black-Box Forgetting (BBF) \cite{kuwana2024black}. We additionally include a learned global suppression baseline that attenuates
all patches uniformly using a CLS-token MLP with identical supervision, while random suppression selects random patches.

\renewcommand{\arraystretch}{1.0}
\begin{table*}[t]
\scriptsize
\setlength{\tabcolsep}{5.2pt}
\centering
\caption{\textbf{Comparison with Baseline CLIP Fine-Tuning Methods.} Average performance over all possible retain-domain and forget-domain combinations for different numbers of forget domains $|D_f|\in\{1,2,3\}$. Since ImageNet contains only two domains, only the setting $|D_f|=1$ is evaluated. Performance is measured using three metrics: \textbf{\textit{Mem}}, classification accuracy on retained domains; \textbf{\textit{For}}, classification error on forget domains; and \textbf{\textit{H}}, the harmonic mean of Mem and For. Higher values indicate better unlearning performance.}
\label{tab:baseline_comparison}
\resizebox{0.96\textwidth}{!}{
\begin{tabular}{c l ccc ccc ccc}
\toprule
\multirow{2}{*}{$|D_{\mathrm{forget}}|$} 
& \multirow{2}{*}{Method}
& \multicolumn{3}{c}{ImageNet}
& \multicolumn{3}{c}{Office-Home}
& \multicolumn{3}{c}{Mini DomainNet} \\
\cmidrule(lr){3-5}\cmidrule(lr){6-8}\cmidrule(lr){9-11}
& & $H \uparrow$ & $Mem \uparrow$ & $For \uparrow$
& $H \uparrow$ & $Mem \uparrow$ & $For \uparrow$
& $H \uparrow$ & $Mem \uparrow$ & $For \uparrow$ \\
\midrule
\multirow{5}{*}{$|D_{\mathrm{forget}}|=1$} 
& LP++ \cite{huang2024lp++}  & 50.69 & 62.22 & 42.76 & 30.46 & 85.06 & 18.55 & 31.73 & 84.72 & 19.52 \\
& CLIPFit \cite{li2024vision} & 71.31 & 63.04 & 82.13 & 43.44 & 83.20 & 29.40 & 53.56 & 84.32 & 39.24 \\
& BBF \cite{kuwana2024black} & 45.56 & 35.92 & 65.91 & 31.25 & 80.94 & 19.74 & 32.12 & 81.80 & 20.03 \\
& ADU \cite{kawamura2026approximate} & 77.02 & 64.13 & 96.73 & 69.96 & 80.93 & 64.34 & 75.56 & 83.32 & 73.06 \\
& \textit{Global Suppression}
& 74.66
& 63.24
& 95.41
& 53.29
& 78.40
& 36.37
& 61.07
& 84.14
& 50.41 \\
& \textit{Random} & 76.18 & 63.58& 97.18 & 59.97 &79.13 & 42.48  & 69.71 & 83.71 & 59.42\\
\rowcolor{red!8}& \textbf{Ours} &78.23  & 64.88 & 98.51 &71.63  &80.21  &66.31  & 80.15 & 82.42 &79.60\\
\midrule

\multirow{5}{*}{$|D_{\mathrm{forget}}|=2$}
& LP++ \cite{huang2024lp++} & $\times$ & $\times$ & $\times$ & 31.11 & 85.54 & 19.01 & 32.21 & 85.29 & 19.85 \\
& CLIPFit \cite{li2024vision} & $\times$ & $\times$ & $\times$ & 40.53 & 83.64 & 26.74 & 51.35 & 84.74 & 36.83 \\
& BBF \cite{kuwana2024black} & $\times$ & $\times$ & $\times$ & 32.54 & 80.42 & 20.34 & 47.95 & 61.48 & 40.52 \\
& ADU \cite{kawamura2026approximate} & $\times$ & $\times$ & $\times$ & 73.58 & 75.61 & 71.89 & 77.03 & 76.51 & 77.66 \\
& \textit{Global Suppression} & $\times$ & $\times$ & $\times$ &55.43  &80.23  &41.23  &61.97  & 85.81 &50.94\\
& \textit{Random} & $\times$ & $\times$ & $\times$  & 64.15& 77.13 & 54.92 & 71.76 & 83.02 &63.19 \\
\rowcolor{red!8}& \textbf{Ours} & $\times$ & $\times$ & $\times$ & 76.50 & 76.32 & 83.17 & 78.02 &76.91  & 79.19 \\
\midrule

\multirow{5}{*}{$|D_{\mathrm{forget}}|=3$}
& LP++ \cite{huang2024lp++} & $\times$ & $\times$ & $\times$ & 33.57 & 86.22 & 20.84 & 34.73 & 85.87 & 21.77 \\
& CLIPFit \cite{li2024vision} & $\times$ & $\times$ & $\times$ & 50.02 & 85.68 & 35.32 & 52.88 & 85.39 & 38.30 \\
& BBF \cite{kuwana2024black} & $\times$ & $\times$ & $\times$ & 29.60 & 78.13 & 26.80 & 39.29 & 80.29 & 26.05 \\
& ADU \cite{kawamura2026approximate} & $\times$ & $\times$ & $\times$ & 75.89 & 72.15 & 80.77 & 79.00 & 75.00 & 83.94 \\
& \textit{Global Suppression} & $\times$ & $\times$ & $\times$ & 59.47 & 80.56 & 47.25 &65.83 &83.77  & 57.02 \\
& \textit{Random} & $\times$ & $\times$ & $\times$  &64.91 & 79.16 & 55.03 & 73.83 & 81.95 & 67.18\\
\rowcolor{red!8} & \textbf{Ours} & $\times$ & $\times$ & $\times$ &76.59  & 71.37 &80.32 &  80.53 &  74.94 & 87.02 \\
\bottomrule
\end{tabular}
}
\vspace{-1.5em}
\end{table*}

\vspace{-1em}
\subsection{Comparison with State-of-the-Art Methods}
\vspace{-0.2em}
Table~\ref{tab:baseline_comparison} summarizes the comparative results. Our method achieves stronger forgetting-retention tradeoffs across given settings. We compare against ADU~\cite{kawamura2026approximate} and a Global Suppression baseline. Under $|D_{\text{forget}}|=1$, our method improves H from 75.56 to 80.15
(+4.59) on Mini DomainNet, from 77.02 to 78.23 (+1.21) on ImageNet, and from
69.96 to 71.63 (+1.67) on Office-Home compared to ADU. Importantly, retain-domain accuracy remains comparable across all three datasets, confirming that the improvement in $H$ is driven by stronger forgetting rather than retention degradation. This demonstrates that selectively suppressing domain-dominant patches removes forget-domain information more precisely than global suppression.
Under $|D_{\text{forget}}|=2$, our method improves $H$ from $73.58$ to $76.50$ ($+2.92$) on Office-Home and from $77.03$ to $78.02$ ($+0.99$) on Mini DomainNet. The consistent improvement across both datasets under a more challenging forgetting setup further supports the effectiveness of patch-level localization. Under $|D_{\text{forget}}|=3$, our method achieves $H=80.53$ on Mini DomainNet ($+1.53$ over ADU). The performance gap over global suppression approaches widens as the number of forget domains increases, suggesting that localized intervention scales better to harder forgetting settings. Global Suppression and random suppression underperform our method across all settings, confirming that uniform patch attenuation suppresses semantically useful representations along with domain-specific ones.

\vspace{-0.8em}
\subsection{Evaluation Under Visually Overlapping Domains}
\vspace{-0.6em}
To the best of our knowledge, there are currently no established multi-domain benchmarks explicitly designed with ambiguous domain boundaries. We therefore analyze Mini DomainNet in Table~\ref{tab:minidomainnet_per_domain}, where domains such as \textit{clipart}, \textit{painting}, and \textit{sketch} exhibit substantial visual overlap. For example, when forgetting the \textit{sketch} domain, its accuracy decreases substantially from 72.54\% to 16.59\%, while visually related domains such as \textit{clipart} decrease from 80.64\% to 79.56\% and \textit{painting} from 78.10\% to 77.91\%, remaining almost unchanged. Similar trends are consistently observed across other forget-domain configurations.

\vspace{-1em}
\subsection{Analysis}
\vspace{-1em}
\begin{table*}[t]
\centering
\begin{minipage}[t]{0.48\linewidth}
\vspace{0pt}
\centering
{\fontsize{6.2pt}{6.7pt}\selectfont
\setlength{\tabcolsep}{3.2pt}
\renewcommand{\arraystretch}{0.88}

\caption{Per-domain accuracy on Mini DomainNet under partially \textbf{\textit{overlapping visual domain characteristics}}. We analyze selective forgetting while preserving performance on similar retain domains.}
\label{tab:minidomainnet_per_domain}

\begin{tabular}{llcccc}
\toprule
Forget Domain & Method & C & P & R & S \\
\midrule
Clipart & ZS-CLIP & 80.64 & 78.10 & 87.94 & 72.54 \\
        & ADU     & 24.76 & 75.40 & 83.97 & 73.97 \\
        & Global Supp. & 24.10 & 71.85 & 79.42 & 68.30 \\
        & Ours    & \cellcolor{red!8} \textbf{15.56} & 76.40 & 84.92 & 72.32 \\
\midrule
Painting & ADU  & 81.59 & 30.32 & 84.92 & 73.97 \\
         & Global Supp. & 74.36 & 29.84 & 78.91 & 69.55 \\
         & Ours & 80.92 & \cellcolor{red!8} \textbf{26.91} & 84.04 & 72.19 \\
\midrule
Real & ADU  & 77.94 & 69.52 & 32.06 & 74.44 \\
     & Global Supp. & 72.48 & 70.62 & 30.71 & 68.94 \\
     & Ours & 77.37 & 69.59 & \cellcolor{red!8} \textbf{24.41} & 73.02 \\
\midrule
Sketch & ADU  & 79.05 & 77.62 & 87.46 & 20.64 \\
       & Global Supp. & 73.15 & 71.08 & 79.36 & 24.92 \\
       & Ours & 79.56 & 77.91 & 86.42 & \cellcolor{red!8} \textbf{16.59} \\
\bottomrule
\end{tabular}
}
\end{minipage}\hfill
\begin{minipage}[t]{0.48\linewidth}
\vspace{0pt}
\centering
\scriptsize
\setlength{\tabcolsep}{3.2pt}
\renewcommand{\arraystretch}{0.95}
\caption{\textbf{Ablation Study.} Ablation results of the domain score and semantic
score components under a single-domain forgetting setting on Benchmark datasets.}
\label{tab:ablation}
\begin{tabular}{lccccc}
\toprule
\multirow{2}{*}{Dataset}
& \multirow{2}{*}{\makecell{Domain\\Score}}
& \multirow{2}{*}{\makecell{Semantic\\Score}}
& \multicolumn{3}{c}{$|D_f|=1$} \\
\cmidrule(lr){4-6}
& & & $H\uparrow$ & $Mem\uparrow$ & $For\uparrow$ \\
\midrule
\multirow{4}{*}{Office-Home}
& -- & -- & 53.29 & 78.40 & 36.37 \\
& \checkmark & -- & 65.95 & 69.43 & 62.80 \\
& -- & \checkmark & 53.51 & 71.92 & 42.60 \\
\rowcolor{red!8} & \checkmark & \checkmark & \textbf{71.63} & \textbf{80.21} & \textbf{66.31} \\
\midrule
\multirow{4}{*}{Mini-DN}
& -- & -- & 52.18 & 85.50 & 39.56 \\
& \checkmark & -- & 70.36 & 82.73 & 61.58 \\
& -- & \checkmark & 66.42 & 85.94 & 54.81 \\
\rowcolor{red!8}& \checkmark & \checkmark & \textbf{80.15} &\textbf{ 82.42} & \textbf{79.60} \\
\bottomrule
\end{tabular}
\end{minipage}
\vspace{-0.7em}
\end{table*}

\vspace{0.3em}
\noindent \underline{\textit{\textbf{Ablation Study.}}} Table~\ref{tab:ablation} presents a component-wise ablation study of the proposed localized domain attenuation framework on Office-Home and Mini DomainNet. We analyze the contribution of domain-aware patch scoring, semantic utility estimation, and their combination through the proposed LDE formulation. Uniform patch suppression yields a limited forgetting-retention tradeoff since all patches are attenuated equally without considering their domain relevance. Incorporating domain-specific patch scores improves forgetting by localizing domain-sensitive regions, while semantic utility estimation helps preserve semantically informative patches. Combining both components through the proposed LDE formulation consistently achieves the better harmonic mean scores across datasets.

\begin{figure}[t]
\centering
\begin{minipage}[t]{0.60\linewidth}
\centering
\includegraphics[width=\linewidth]{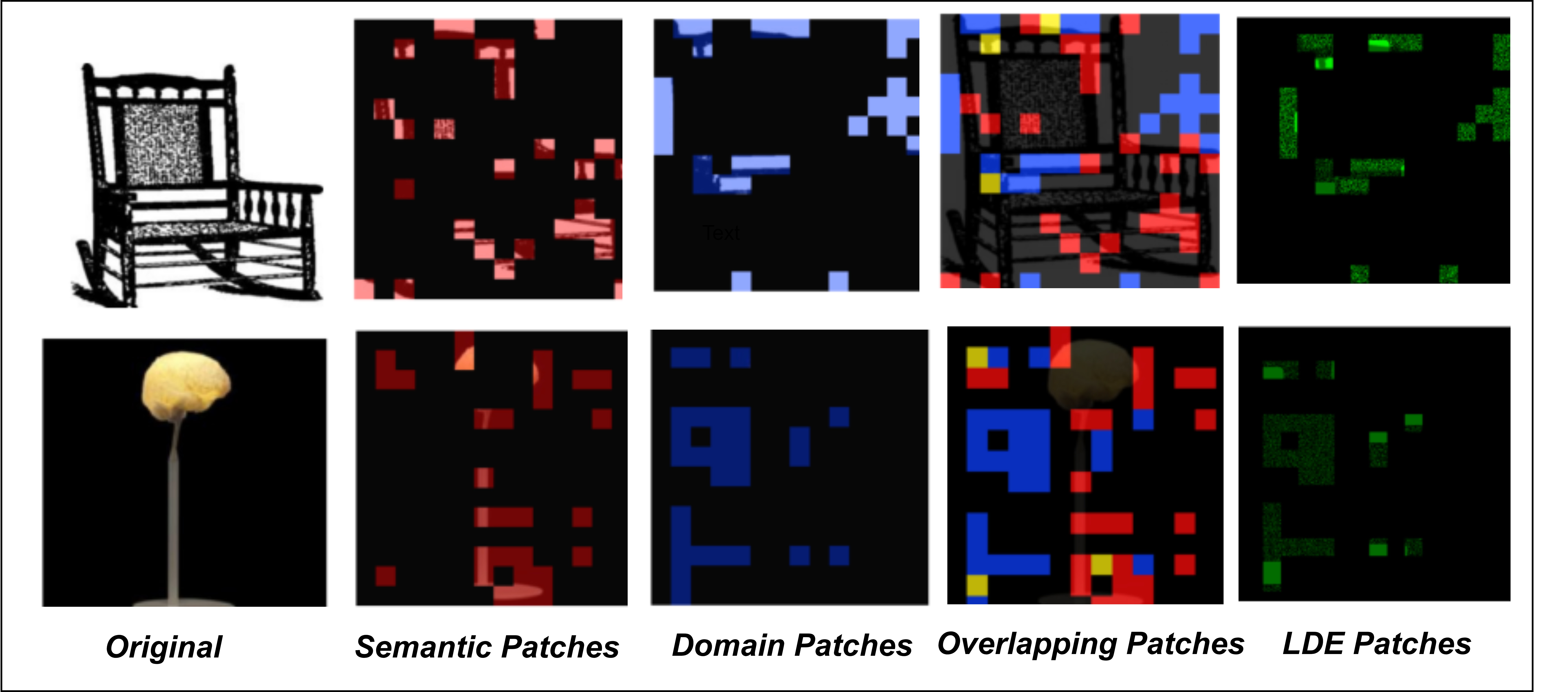}
\end{minipage}
\hfill
\begin{minipage}[t]{0.39\linewidth}
\centering
\includegraphics[width=\linewidth]{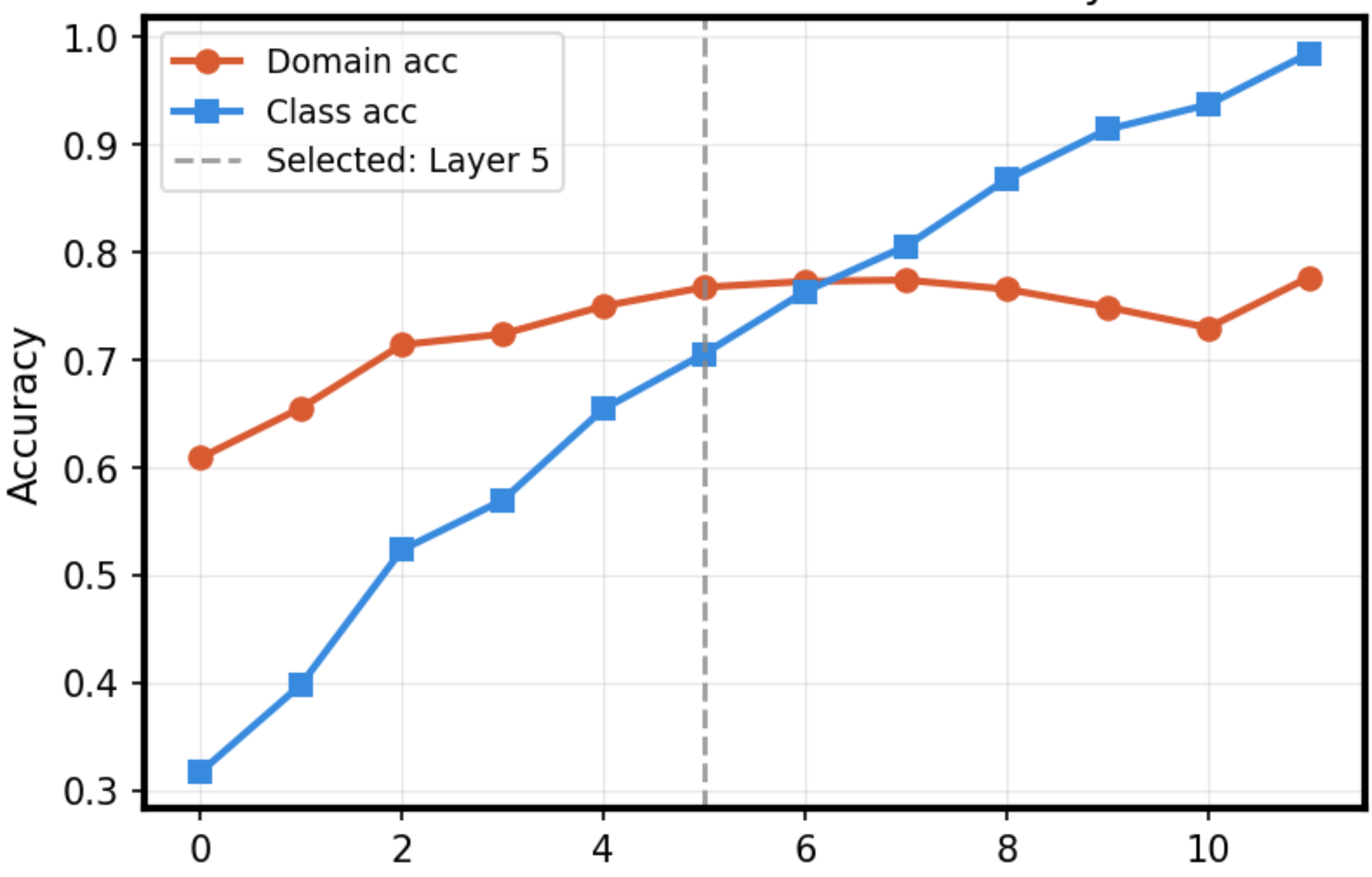}
\end{minipage}
\vspace{-1.7em}
\caption{\textbf{Ablation analysis and visualization of localized patch effects.} \textit{\textbf{Left}}: patch-level visualization of semantic utility (blue), domain sensitivity (red), overlapping regions (yellow), and the resulting Localized Domain Effect (LDE) maps (green) across representative samples. \textit{\textbf{Right}}: domain and class probe accuracy across transformer layers of the frozen model, with the selected intervention layer marked at layer $l=5$.}
\label{fig:combined_ablation}
\vspace{-1em}
\end{figure}

\vspace{0.2em}
\noindent \underline{\textit{\textbf{Visualization of Localized Domain Effect.}}} Fig.~\ref{fig:combined_ablation} (a) visualizes the estimated domain sensitivity, semantic utility, and resulting Localized Domain Effect (LDE) maps across different samples. The domain sensitivity maps primarily highlight domain-dependent regions such as texture, background, illumination, and style patterns, whereas semantic utility maps concentrate more strongly around object-centric structures. The resulting LDE maps selectively emphasize regions whose forget-domain contribution is disproportionately higher than their semantic utility, demonstrating that domain-related information is spatially localized rather than uniformly distributed across patch representations. These observations support the effectiveness of localized patch-level suppression for approximate domain unlearning.

\begin{figure}[t]
\centering
\includegraphics[width=0.82\linewidth]{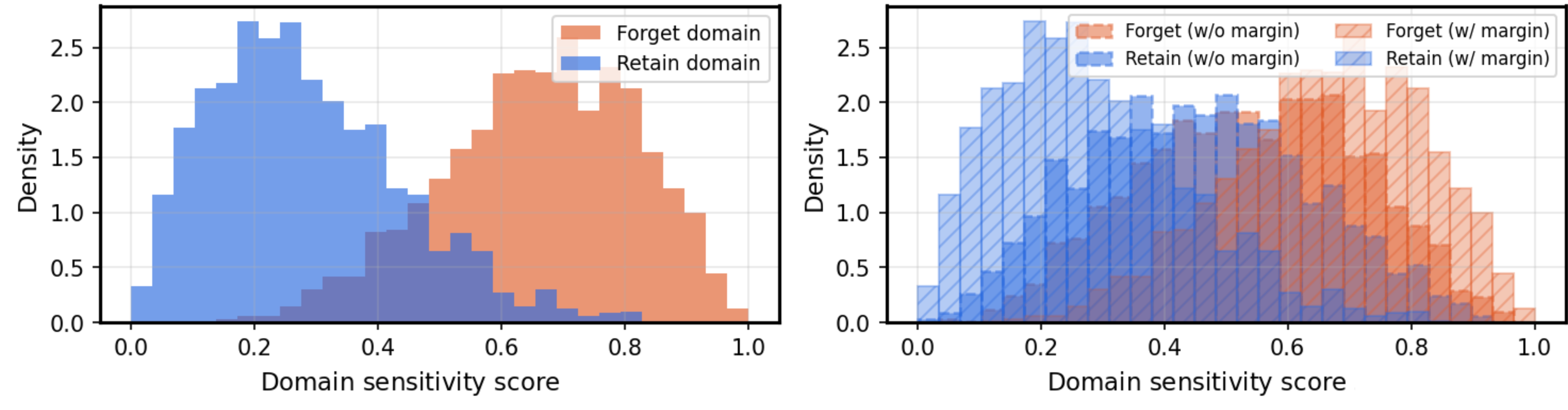}
\vspace{-0.9em}
\caption{\textbf{Patch-domain scorer analysis.}
(a) Distribution of patch-domain sensitivity scores for forget-domain and retained-domain patches.
(b) Effect of the margin objective on score separation between forget-domain and retained-domain patches.}
\label{fig:patch_score_analysis}
\vspace{-1.3em}
\end{figure}

\vspace{0.5em}
\noindent \underline{\textit{\textbf{Analysis of Patch Domain Scorer}}}
We analyze the proposed patch-domain scorer to verify whether it successfully localizes domain-sensitive visual regions. Fig.~\ref{fig:patch_score_analysis}(a) shows a clear separation between forget-domain and retained-domain patch distributions, where forget-domain patches consistently receive higher domain sensitivity scores despite training with only image-level weak supervision. Fig.~\ref{fig:patch_score_analysis} further shows that incorporating the margin objective significantly improves this separation by pushing forget-domain patches toward higher scores while suppressing retained-domain patches toward lower scores. These observations suggest that the proposed scorer learns meaningful localized domain structure rather than assigning arbitrary patch importance, enabling more selective patch-level unlearning.

\vspace{0.5em}
\noindent \underline{\textit{\textbf{Effect of Intermediate Transformer Layer Selection. }}} Following \cite{alain2016understanding}, we train linear domain and class probes across all transformer layers of frozen CLIP to identify where domain evidence is accessible without interference from consolidated class representations. As shown in Fig.~\ref{fig:combined_ablation}(b), domain probe accuracy exceeds class probe accuracy through Layer 5 (0.76 vs.\ 0.70), after which class accuracy rises steeply and the gap collapses. We therefore select Layer 5 as the intervention point: it is the deepest layer at which domain representations remain linearly separable while class consolidation is incomplete, satisfying the selective intervention criterion of \cite{gandikota2023erasing}. Detailed analysis is provided in the \textit{Supplementary}.

\vspace{1em}
\begin{wraptable}{r}{0.45\linewidth}
\vspace{-0.8em}
\centering
\scriptsize
\renewcommand{\arraystretch}{0.95}
\setlength{\tabcolsep}{5pt}
\caption{\textbf{Patch granularity ablation.}}
\vspace{0.4em}
\label{tab:patch_granularity}
\begin{tabular}{lccc}
\toprule
\textbf{Patch} & \textbf{H}$\uparrow$ & \textbf{Mem}$\uparrow$ & \textbf{For}$\uparrow$ \\
\midrule
$3{\times}3$    & 69.36 & 74.82 & 64.69 \\
$7{\times}7$    & 71.10 & 75.12 & \textbf{67.50} \\
\rowcolor{red!8}
$14{\times}14$  & \textbf{71.63} & 80.21 & 66.31 \\
$28{\times}28$  & 70.84 & \textbf{81.34} & 62.60 \\
\bottomrule
\end{tabular}
\vspace{-0.7em}
\end{wraptable}
\noindent \underline{\textit{\textbf{Sensitivity to Patch Granularity.}}}
We analyze the sensitivity of the proposed framework to patch granularity by varying the spatial partition size used for patch-level intervention. As shown in Table~\ref{tab:patch_granularity}, the retain--forget tradeoff remains stable across different granularities, indicating that the method is not highly sensitive to the precise patch partition size. Intermediate granularity provides the best \textit{$H$} with patch size $14{\times}14$.

\vspace{-0.5em}
\section{Discussion and Conclusion}
\vspace{-0.6em}
We introduced a localized patch-guided framework for approximate domain unlearning in
vision-language models by exploiting the spatial structure of vision transformers. We find that
certain patches contribute disproportionately more to forget-domain prediction than to semantic
recognition. Motivated by this observation, our framework estimates the relative domain and
semantic contribution of each patch and selectively suppresses forget-domain-dominant regions
prior to global aggregation. Extensive experiments demonstrate stronger forgetting-retention
tradeoffs through localized intervention. A limitation arises when domain and semantic evidence
strongly overlap within the same tokens, since semantic calibration may preserve such regions.
Resolving this overlap remains an important direction for approximate domain unlearning.
\bibliography{egbib}
\end{document}


\maketitle

\begin{abstract}
This \textit{supplementary} material provides additional experimental analysis and implementation details for the proposed localized patch-guided domain unlearning framework. We first present the complete algorithm of the proposed method, followed by additional results on large-scale multi-domain benchmarks. We further analyze forgetting behavior under visually overlapping domains and examine whether localized patch suppression remains selective without globally degrading semantic representations. We additionally study the computational efficiency of the proposed framework and investigate the behavior of the patch-domain scorer under different intervention settings. Finally, we provide detailed implementation details, hyperparameter configurations, and dataset descriptions used throughout all experiments.
\end{abstract}

\section{Algorithm}
The proposed localized patch-guided domain unlearning framework selectively suppresses domain-sensitive patch representations while preserving semantically informative regions within the feature space of a vision transformer. For each patch token, a lightweight patch-domain scorer estimates domain sensitivity, while semantic utility is computed using cosine alignment with retained-class text embeddings. These quantities are combined to identify patches whose contribution is disproportionately dominated by forget-domain information. Only highly domain-dominant patches are attenuated, whereas semantically useful patches are preserved. The modified patch representations are then propagated through the remaining transformer layers for final prediction. The complete procedure is summarized in Algorithm~\ref{alg:ours_compact}.
\begin{algorithm}[t]
\caption{\textbf{\textit{Localized Patch-Guided Domain Unlearning}}}
\label{alg:ours_compact}
\small

\begin{algorithmic}[1]

\Require Image $x$, ViT encoder $\phi$, forget-domain scorer $g(\cdot)$, retain-class text embeddings $\{t_c\}_{c=1}^{C}$, threshold $\tau$, attenuation strength $\alpha$

\Ensure Modified patch sequence $\widetilde{H}^l(x)$

\State Extract intermediate patch tokens
$
H^l(x)=\{h_i^l\}_{i=1}^{N}
$

\For{each patch token $h_i^l$}

    \State Compute domain sensitivity score
    $
    s_i=\sigma(g(h_i^l))
    $

    \State Compute semantic utility score
    $
    u_i=\max_c \cos(h_i^l,t_c)
    $

    \State Compute localized domain effect
    $
    r_i=\frac{s_i}{u_i+\epsilon}
    $

    \If{$r_i>\tau$}

        \State Suppress domain-dominant patch
        $
        \widetilde{h}_i^l
        \leftarrow
        (1-\alpha \bar r_i)h_i^l
        $

    \Else

        \State Preserve semantically useful patch
        $
        \widetilde{h}_i^l
        \leftarrow
        h_i^l
        $

    \EndIf

\EndFor

\State Form modified patch sequence
$
\widetilde{H}^l(x)=
\{\widetilde{h}_1^l,\ldots,\widetilde{h}_N^l\}
$

\State Forward $\widetilde{H}^l(x)$ through the remaining transformer layers

\Return $\widetilde{H}^l(x)$

\end{algorithmic}
\end{algorithm}

\section{Results on DomainNet}
We additionally evaluate our method on DomainNet~\cite{peng2019moment}, a large-scale multi-domain benchmark containing substantial visual diversity across domains. DomainNet consists of six domains, namely \textit{Clipart}, \textit{Infograph}, \textit{Painting}, \textit{Quickdraw}, \textit{Real}, and \textit{Sketch}, spanning 345 object categories. Compared to Office-Home and Mini-DomainNet, DomainNet exhibits significantly larger domain variability and more severe distribution shifts, making selective domain suppression substantially more challenging. The quantitative results are reported in Table~\ref{tab:domainnet_results}. Our localized patch-guided suppression framework consistently outperforms the baseline approach across all forgetting settings, achieving substantially stronger forgetting-retention trade-offs. These results further demonstrate that localized suppression remains effective even under large-scale and highly diverse domain distributions.
\begin{table*}[t]
\centering
\small
\setlength{\tabcolsep}{6pt}
\renewcommand{\arraystretch}{1.12}
\caption{\textbf{Results on DomainNet.}}
\label{tab:domainnet_results}
\begin{tabular}{lccccccccc}
\hline
\multirow{2}{*}{Method}
& \multicolumn{3}{c}{$|D_{\mathrm{forget}}|=1$}
& \multicolumn{3}{c}{$|D_{\mathrm{forget}}|=2$}
& \multicolumn{3}{c}{$|D_{\mathrm{forget}}|=3$} \\
\cline{2-10}
& $H\uparrow$ & $Mem\uparrow$ & $For\uparrow$
& $H\uparrow$ & $Mem\uparrow$ & $For\uparrow$
& $H\uparrow$ & $Mem\uparrow$ & $For\uparrow$ \\
\hline
Zero-shot CLIP
& 48.26 & 56.91 &41.83
& 36.85 & 53.28 & 28.16
& 36.85 & 53.28 & 28.16 \\
ADU \cite{kawamura2026approximate}
& 66.81 & 58.86 & 77.23
& 67.81 & 58.89 & 79.90
& 68.53 & 59.06 & 82.48 \\
\rowcolor{red!8}
Ours
& 69.73 & 59.17 & 84.36
& 68.92 & 56.19 & 89.24
&  68.78 & 57.08 &86.49  \\
\hline
\end{tabular}
\end{table*}

\FloatBarrier
\section{Generalization to Unseen Domains}
We additionally evaluate the proposed framework under unseen-domain settings to examine whether localized patch-guided suppression preserves transferable semantic representations beyond the domains explicitly observed during optimization. Existing approximate domain unlearning methods are primarily evaluated only on retained and forget domains used during training. However, in realistic deployment scenarios, vision-language models may encounter previously unseen domains whose visual statistics partially overlap with both retained and forget domains. Excessive global perturbation may therefore unintentionally damage domain-invariant semantic representations and degrade transferability under distribution shift.

To analyze this behavior, we construct unseen-domain evaluation settings in which one domain is excluded during optimization and used only at inference time. This setup allows us to evaluate whether the proposed localized suppression framework removes domain-specific evidence while preserving semantically transferable representations. In particular, if the method merely suppresses global visual features, performance on unseen domains would deteriorate substantially due to excessive perturbation of shared semantic structure. Conversely, preserving strong unseen-domain performance would suggest that the proposed method selectively attenuates localized forget-domain cues while retaining domain-invariant semantic information.

The quantitative results demonstrate that localized patch-guided suppression consistently maintains stronger recognition performance on unseen domains compared to global suppression baselines. These observations suggest that selective patch-level intervention perturbs domain-specific evidence more precisely than uniform global suppression, leading to improved generalization under domain shift.

\begin{table}[t]
\centering
\small
\setlength{\tabcolsep}{9pt}
\renewcommand{\arraystretch}{1.12}
\caption{\textbf{Generalization to Unseen Domains on Office-Home.}
Unseen-domain accuracy is reported separately and is not included in $H$.}
\label{tab:unseen_domain_officehome}
\begin{tabular}{lcccc}
\hline
\textbf{Method} & \textbf{$H \uparrow$} & \textbf{Mem $\uparrow$} & \textbf{For $\uparrow$} & \textbf{Unseen $\uparrow$} \\
\hline
Baseline & 61.28 & 72.84 & 53.01 & 55.62 \\
ADU~\cite{kawamura2026approximate} & 67.14 & 76.91 & 59.83 & 61.48 \\
\rowcolor{red!8}
Ours & \textbf{73.42} & \textbf{81.37} & \textbf{66.89} & \textbf{72.15} \\
\hline
\end{tabular}
\end{table}

\begin{table}[t]
\centering
\small
\setlength{\tabcolsep}{9pt}
\renewcommand{\arraystretch}{1.12}
\caption{\textbf{Generalization to Unseen Domains on DomainNet.}The results evaluate whether forgetting on the target domain generalizes while preserving performance on retained and previously unseen domains under distribution shift. Unseen-domain accuracy is reported separately and is not included in $H$.}
\label{tab:unseen_domain}
\begin{tabular}{lcccc}
\hline
\textbf{Method} & \textbf{$H \uparrow$} & \textbf{Mem $\uparrow$} & \textbf{For $\uparrow$} & \textbf{Unseen $\uparrow$} \\
\hline
Baseline & 58.42 & 69.31 & 50.48 & 46.72 \\
ADU~\cite{kawamura2026approximate} & 64.83 & 74.26 & 57.43 & 53.91 \\
\rowcolor{red!8}
Ours & \textbf{71.54} & \textbf{80.18} & \textbf{64.82} & \textbf{67.35} \\
\hline
\end{tabular}
\end{table}

\section{Per-Domain Accuracy}
We report per-domain accuracy on Office-Home, Mini-DomainNet (in the main paper), and DomainNet to evaluate whether our localized patch-guided suppression framework performs selective domain forgetting without damaging visually related retained domains. Several domains share substantial visual similarity (e.g., Clipart vs.\ Painting and Sketch vs.\ Quickdraw), making selective suppression particularly challenging.The results show that our method consistently reduces accuracy on the forgotten domain while preserving performance on the remaining domains. This behavior indicates that the proposed patch-level suppression mechanism selectively attenuates domain-dominant visual cues instead of globally degrading the representation space.

\begin{table*}[t]
\centering
\small
\setlength{\tabcolsep}{10pt}
\renewcommand{\arraystretch}{1.12}
\caption{\textbf{Per-Domain Accuracy on Office-Home.}}
\label{tab:officehome_per_domain}
\begin{tabular}{lcccc}
\hline
Method & \textit{art} & \textit{clipart} & \textit{product} & \textit{real} \\
\hline
Zero-shot CLIP & 74.34 & 60.97 & 80.43 & 81.29 \\
Ours ($D_{\mathrm{forget}}=\{\textit{art}\}$) 
& \cellcolor{red!8}\textbf{31.94} & 69.19 & 88.94 & 76.46 \\
Ours ($D_{\mathrm{forget}}=\{\textit{clipart}\}$) 
& 76.64 & \cellcolor{red!8}\textbf{12.64} & 88.19 & 81.93 \\
Ours ($D_{\mathrm{forget}}=\{\textit{product}\}$) 
& 80.38 & 72.15 & \cellcolor{red!8}\textbf{32.77} & 77.85 \\
Ours ($D_{\mathrm{forget}}=\{\textit{real}\}$) 
& 66.19 & 69.71 & 79.44 & \cellcolor{red!8}\textbf{47.85} \\
\hline
\end{tabular}
\end{table*}
\begin{table*}[t]
\centering
\scriptsize
\setlength{\tabcolsep}{8.5pt}
\renewcommand{\arraystretch}{1.05}
\caption{\textbf{Per-Domain Accuracy on DomainNet.}}
\label{tab:domainnet_per_domain}
\begin{tabular}{lcccccc}
\hline
Method & \textit{clipart} & \textit{infograph} & \textit{painting} & \textit{quickdraw} & \textit{real} & \textit{sketch} \\
\hline
Zero-shot CLIP & 71.84 & 50.01 & 65.36 & 14.91 & 83.42 & 63.07 \\
Ours ($D_{\mathrm{forget}}=\{\textit{clipart}\}$) 
& \cellcolor{red!8}\textbf{26.30} & 52.27 & 65.42 & 29.69 & 79.80 & 63.26 \\
Ours ($D_{\mathrm{forget}}=\{\textit{infograph}\}$) 
& 70.93 & \cellcolor{red!8}\textbf{9.46} & 62.57 & 34.94 & 79.50 & 61.25 \\
Ours ($D_{\mathrm{forget}}=\{\textit{painting}\}$) 
& 72.59 & 50.11 & \cellcolor{red!8}\textbf{19.57} & 24.64 & 75.87 & 63.00 \\
Ours ($D_{\mathrm{forget}}=\{\textit{quickdraw}\}$) 
& 74.91 & 50.33 & 67.63 & \cellcolor{red!8}\textbf{5.02} & 80.43 & 66.47 \\
Ours ($D_{\mathrm{forget}}=\{\textit{real}\}$) 
& 70.61 & 52.40 & 65.72 & 28.96 & \cellcolor{red!8}\textbf{30.57} & 63.29 \\
Ours ($D_{\mathrm{forget}}=\{\textit{sketch}\}$) 
& 71.42 & 51.63 & 67.20 & 30.32 & 80.87 & \cellcolor{red!8}\textbf{22.78} \\
\hline
\end{tabular}
\end{table*}
\begin{table}[!t]
\centering
\caption{\textbf{Robustness across VLM architectures and model scales on
Office-Home under $|D_f|=1$.}
We compare the proposed method with ADU~\cite{kawamura2026approximate}
under CLIP ViT-B/16, the larger CLIP ViT-L/14 backbone, and
SigLIP~\cite{zhai2023sigmoid}.}
\label{tab:vlm_robustness}

\small
\setlength{\tabcolsep}{4.0pt}
\renewcommand{\arraystretch}{0.95}

\begin{tabular}{llcccc}
\toprule
\textbf{VLM / Backbone} &
\textbf{Method} &
$H\uparrow$ &
Mem$\uparrow$ &
For$\uparrow$ &
$\Delta H$ \\
\midrule

\multirow{2}{*}{CLIP ViT-B/16}
& ADU~\cite{kawamura2026approximate}
& 69.96 & \textbf{80.93} & 64.34 & -- \\
& Ours
& \textbf{71.63} & 80.21 & \textbf{66.31} & \textbf{+1.67} \\

\midrule

\multirow{2}{*}{CLIP ViT-L/14}
& ADU~\cite{kawamura2026approximate}
& 72.16 & \textbf{83.28} & 66.72 & -- \\
& Ours
& \textbf{74.24} & 82.87 & \textbf{70.18} & \textbf{+2.08} \\

\midrule

\multirow{2}{*}{SigLIP~\cite{zhai2023sigmoid}}
& ADU~\cite{kawamura2026approximate}
& 72.36 & \textbf{83.46} & 66.85 & -- \\
& Ours
& \textbf{74.55} & 83.02 & \textbf{70.48} & \textbf{+2.19} \\

\bottomrule
\end{tabular}
\end{table}

\section{Robustness Across VLM Architectures and Model Scales.}
To evaluate whether the proposed localized intervention generalizes beyond
CLIP ViT-B/16, we additionally consider the larger CLIP ViT-L/14 backbone
and SigLIP~\cite{zhai2023sigmoid}. As shown in
Table~\ref{tab:vlm_robustness}, the proposed method consistently improves
the forgetting--retention harmonic mean over ADU~\cite{kawamura2026approximate}
across both increased model scale and a distinct VLM architecture. The gains
are primarily driven by stronger forgetting while retain-domain accuracy
remains close to the corresponding ADU baseline, consistent with the behavior
observed with CLIP ViT-B/16.
\section{Computational Complexity}
Table~\ref{tab:complexity} summarizes GPU memory usage and training time on Office-Home using a single NVIDIA H100/H200 GPU. Compared with ADU~\cite{kawamura2026approximate}, our localized patch-guided suppression framework requires substantially lower computational overhead while achieving stronger forgetting-retention trade-offs. Since our method operates using a lightweight patch-domain scorer with a frozen CLIP backbone, it avoids the expensive prompt optimization, domain disentanglement objectives, and instance-wise prompt generation modules used in ADU. Furthermore, incorporating semantic utility estimation introduces only marginal additional computational cost, increasing memory usage by less than 1 GB and training time by only a few seconds. These results demonstrate that the proposed localized suppression strategy remains computationally efficient and practical for large-scale vision-language models.
\begin{table}[H]
\centering
\small
\setlength{\tabcolsep}{7pt}
\renewcommand{\arraystretch}{1.12}
\caption{\textbf{Computational Complexity on Office-Home.}}
\label{tab:complexity}

\begin{tabular}{lcc}
\hline
Method & Memory [GB] & Time [s] \\
\hline
ADU~\cite{kawamura2026approximate}
& 10.7 & 550 \\
Global Suppression
& 6.8 & 214 \\
Ours w/o Semantic Utility
& 7.1 & 236 \\
Ours
& 7.4 & 248 \\
\hline
\end{tabular}
\end{table}

\section{Sensitivity to Hyperparameters } 
We analyze the sensitivity of our method to the attenuation strength $\alpha$ and the margin-loss weight $\lambda$. As shown in Fig.~\ref{fig:hyperparameter_layer_analysis}\textit{(a,b)}, performance improves sharply as $\alpha$ increases from $0.5$ to $1.0$, peaks around $\alpha=1.5$, and then remains stable up to $\alpha=5.0$, indicating that the proposed patch-level attenuation is not overly sensitive once sufficient attenuation is applied. For $\lambda$, performance peaks at a moderate value of $\lambda=0.3$; larger values gradually reduce $H$, suggesting that excessive margin weighting can over-constrain noisy patch-level supervision. Overall, the method demonstrates stable behavior across a wide hyperparameter range. We also evaluate the LDE threshold $\tau$ in Fig.~\ref{fig:hyperparameter_layer_analysis}\textit{ (c)}, which controls when a patch is considered sufficiently domain-dominant for attenuation. Performance remains stable around $\tau=0.3$, supporting our default choice where attenuation is applied only when forget-domain contribution exceeds semantic utility.

\begin{figure*}[t]
\centering

\includegraphics[height=3.2cm,width=0.32\textwidth]{Figures/alpha.png}
\hfill
\includegraphics[height=3.2cm,width=0.32\textwidth]{Figures/lambda_plot_clean.png}
\hfill
\includegraphics[height=3.2cm,width=0.32\textwidth]{Figures/tau_rectangular.png}

\vspace{-0.5em}

\begin{minipage}{0.32\textwidth}
\centering
{\scriptsize \textbf{(a) Attenuation strength $\alpha$}}
\end{minipage}
\hfill
\begin{minipage}{0.32\textwidth}
\centering
{\scriptsize \textbf{(b) Margin weight $\lambda$}}
\end{minipage}
\hfill
\begin{minipage}{0.32\textwidth}
\centering
{\scriptsize \textbf{(c) Threshold weight $\tau$}}
\end{minipage}

\vspace{-0.4em}

\caption{\textbf{Additional analysis.}
(a) Sensitivity to the attenuation strength $\alpha$.
(b) Sensitivity to the margin-loss weight $\lambda$.
(c) Sensitivity to the threshold weight $\tau$.}
\label{fig:hyperparameter_layer_analysis}

\vspace{-0.8em}
\end{figure*}

\begin{wraptable}{r}{0.46\textwidth}
\vspace{-0.8em}
\centering
\caption{\textbf{Effect of varying the fraction of highest-LDE patch tokens
attenuated on Office-Home under $|D_f|=1$.}}
\label{tab:attenuation_budget}
\scriptsize
\setlength{\tabcolsep}{3.2pt}
\renewcommand{\arraystretch}{0.92}
\begin{tabular}{cccc}
\toprule
Tokens (\%) & $H\uparrow$ & Mem$\uparrow$ & For$\uparrow$ \\
\midrule
5  & 68.12 & \textbf{80.44} & 59.31 \\
10 & 70.54 & 80.31 & 63.33 \\
20 & \textbf{71.63} & 80.15 & 66.57 \\
30 & 71.30 & 79.48 & 67.02 \\
40 & 70.46 & 78.46 & 67.35 \\
50 & 69.41 & 77.18 & \textbf{67.95} \\
\bottomrule
\end{tabular}
\vspace{-0.7em}
\end{wraptable}

\noindent\textbf{Sensitivity to Attenuation Budget.}
We further analyze sensitivity to the fraction of highest-LDE patch tokens
selected for attenuation on Office-Home under $|D_f|=1$. As shown in
Table~\ref{tab:attenuation_budget}, increasing the attenuation budget improves
forgetting, while excessive attenuation gradually reduces retained-domain
performance. The harmonic mean is highest at 20\%, indicating a favorable
forgetting--retention trade-off across intermediate attenuation budgets.

\begin{wraptable}{r}{0.46\textwidth}
\vspace{-0.8em}
\centering
\caption{\textbf{Effect of the number of labelled images per domain used to
train the patch-domain scorer on Office-Home.}}
\label{tab:scorer_samples}
\scriptsize
\setlength{\tabcolsep}{3.2pt}
\renewcommand{\arraystretch}{0.92}
\begin{tabular}{cccc}
\toprule
Images & $H\uparrow$ & Mem$\uparrow$ & For$\uparrow$ \\
\midrule
2  & 67.84 & 79.62 & 58.95 \\
4  & 70.12 & 79.88 & 62.50 \\
8  & 71.63 & 80.21 & 66.31 \\
16 & 72.18 & 80.34 & 66.57 \\
32 & 72.82 & 80.39 & 66.82 \\
\bottomrule
\end{tabular}
\vspace{-0.7em}
\end{wraptable}

\noindent\textbf{Sensitivity to Scorer Supervision.}
We additionally evaluate sensitivity to the number of labelled images per
domain used to train the patch-domain scorer. Table~\ref{tab:scorer_samples}
shows that performance improves progressively with additional supervision.
Using only eight labelled images per domain already yields $H=71.63$, while
increasing supervision to 32 images provides a modest improvement to $H=72.82$,
indicating that the scorer remains effective under limited supervision.

\begin{figure}[h]
    \centering
    \includegraphics[width=0.9\textwidth]{Figures/plot_scatter_all_colored.png}
    \caption{\textbf{Layer-wise domain--class patch disentanglement in CLIP model.}
    Per-patch domain and class probe scores across transformer layers. Early layers exhibit strong spatial separation between domain-dominant (red) and class-dominant (blue) patches, while deeper layers progressively entangle both representations. Layer 5 (orange border) is the deepest layer where meaningful spatial disentanglement is preserved, motivating its selection for localized patch suppression.}
    \label{fig:scatter}
\end{figure}
\begin{table}[t]
\centering
\scriptsize
\setlength{\tabcolsep}{9pt}
\renewcommand{\arraystretch}{0.92}

\caption{\textbf{Effect of Intervention Layer Selection.}}
\label{tab:layer_analysis}

\vspace{-0.3em}

\begin{tabular}{c|ccc}
\toprule

\textbf{Layer $l$}
& \textbf{$H \uparrow$}
& \textbf{$Mem \uparrow$}
& \textbf{$For \uparrow$} \\

\midrule

0  & 52.1 & 71.3 & 40.8 \\
1  & 57.3 & 73.1 & 45.2 \\
2  & 62.8 & 75.4 & 53.6 \\
3  & 66.4 & 77.2 & 57.9 \\
4  & 69.2 & 78.8 & 61.4 \\

\rowcolor{red!8}
\textbf{5}
& \textbf{71.6}
& \textbf{80.2}
& \textbf{66.3} \\

6  & 69.1 & 79.1 & 61.2 \\
7  & 65.3 & 77.4 & 56.1 \\
8  & 60.2 & 75.2 & 49.8 \\
9  & 55.8 & 73.1 & 44.3 \\
10 & 51.4 & 71.8 & 39.7 \\
11 & 47.9 & 70.2 & 35.1 \\

\bottomrule
\end{tabular}

\vspace{-0.5em}

\end{table}

\section{Effect of Layer Selection}
We analyze the effect of intervention layer selection by applying localized
patch suppression at different transformer layers of the CLIP ViT-B/16 visual
encoder. To characterize how domain-sensitive and semantic information evolve
across transformer depth, we examine the per-patch domain and class probe
scores at each transformer layer. As shown in
Fig.~\ref{fig:scatter}, early and intermediate layers exhibit clearer spatial
separation between domain-dominant and class-dominant patches, whereas deeper
layers show progressively greater overlap between the two types of evidence.
Notably, Layer 5 retains meaningful separation between domain-sensitive and
semantically informative patches, providing a suitable representation for
selective patch-level intervention.

The corresponding downstream unlearning results are reported in
Table~\ref{tab:layer_analysis}. Performance improves progressively from
shallow to intermediate layers, with the harmonic mean increasing from
$H=52.1$ at Layer 0 to $H=71.6$ at Layer 5. Beyond Layer 5, performance
decreases consistently, reaching $H=47.9$ at Layer 11. This behavior indicates
that shallow representations provide insufficiently discriminative domain
information, whereas interventions at deeper layers increasingly affect
semantic representations. Layer 5 provides the best forgetting--retention
trade-off, achieving $H=71.6$, Mem$=80.2$, and For$=66.3$.

\begin{wraptable}{r}{0.46\textwidth}
\vspace{-0.8em}
\centering
\caption{\textbf{Domain--semantic token overlap,} measured as the mean
percentage of shared patch-token indices between matched top-$q\%$
domain-sensitivity and semantic-utility sets over the evaluation set.}
\label{tab:token_overlap}
\scriptsize
\setlength{\tabcolsep}{4.0pt}
\renewcommand{\arraystretch}{0.92}
\begin{tabular}{cccc}
\toprule
Budget & OH & Mini-DN & DN \\
\midrule
5\%  & 15.82 & 13.94 & 12.71 \\
10\% & 24.63 & 21.47 & 19.84 \\
20\% & 36.81 & 33.92 & 31.56 \\
\bottomrule
\end{tabular}
\vspace{-0.7em}
\end{wraptable}

\vspace{0.3cm}
\noindent\textbf{Domain--Semantic Token Overlap.}
To complement the layer-wise probe analysis, we quantify the spatial overlap
between domain-sensitive and semantically informative patch sets.
Table~\ref{tab:token_overlap} reports the mean percentage of shared patch-token
indices between matched top-$q\%$ domain-sensitivity and semantic-utility sets.
Across Office-Home, Mini-DomainNet, and DomainNet, the overlap remains partial
even as the selection budget increases. This provides dataset-level evidence
that domain-sensitive and semantic evidence are not uniformly co-localized,
supporting localized intervention before stronger semantic consolidation at
deeper transformer layers.

\section{Additional Experimental Details}
\subsection{Dataset Details}
We evaluate our method on four multi-domain visual recognition benchmarks commonly used in domain adaptation and domain generalization research: Office-Home~\cite{venkateswara2017deep}, Mini-DomainNet~\cite{zhou2021domain}, DomainNet~\cite{peng2019moment}, and ImageNet-Sketch~\cite{wang2019learning}.

\noindent\textbf{Office-Home.}
Office-Home~\cite{venkateswara2017deep} contains approximately 15.5K images spanning 65 object categories across four visually distinct domains: \textit{Art}, \textit{Clipart}, \textit{Product}, and \textit{Real-World}. The benchmark exhibits substantial appearance variation across domains while preserving shared semantic categories, making it suitable for evaluating domain-specific forgetting under moderate distribution shift.

\noindent\textbf{Mini-DomainNet.}
Mini-DomainNet~\cite{zhou2021domain} is a reduced variant of DomainNet designed for efficient multi-domain evaluation. It contains 126 classes distributed across four domains: \textit{Clipart}, \textit{Painting}, \textit{Real}, and \textit{Sketch}. Compared to Office-Home, Mini-DomainNet exhibits stronger stylistic variation and larger inter-domain discrepancies, providing a more challenging benchmark for selective domain suppression.

\noindent\textbf{DomainNet.}DomainNet~\cite{peng2019moment} is a large-scale multi-domain benchmark containing approximately 0.6 million images from 345 categories across six domains: \textit{Clipart}, \textit{Infograph}, \textit{Painting}, \textit{Quickdraw}, \textit{Real}, and \textit{Sketch}. The benchmark contains substantial domain diversity ranging from natural images to highly abstract drawings, making it particularly suitable for analyzing domain-sensitive representations in vision-language models.

\noindent\textbf{ImageNet-Sketch. } Following prior approximate domain unlearning work~\cite{kawamura2026approximate}, we additionally evaluate on the ImageNet/ImageNet-Sketch setting, where ImageNet-Sketch~\cite{wang2019learning} serves as a shifted sketch-style domain corresponding to the original ImageNet classes~\cite{deng2009imagenet}. This benchmark evaluates whether the proposed localized suppression mechanism can selectively reduce sketch-domain behavior while preserving performance on the original natural-image domain.

\subsection{Implementation Details}

We implement our method using a CLIP-based vision-language model with a ViT-B/16 visual encoder. For an input image, the visual encoder produces a sequence of patch tokens, and we apply localized suppression at an intermediate transformer layer. Unless otherwise stated, we use the patch tokens before the selected transformer block as the intervention representation. For ViT-B/16 with $224\times224$ input resolution, this gives $14\times14=196$ patch tokens per image.

\noindent\textbf{Patch-domain scorer.}
We use a lightweight two-layer MLP
$[\mathrm{Linear}(768\!\rightarrow\!128)\rightarrow\mathrm{GELU}
\rightarrow\mathrm{Linear}(128\!\rightarrow\!1)]$
as the patch-domain scorer to distinguish forget-domain patches from
retained-domain patches. Since only image-level domain labels are available,
all patch tokens extracted from forget-domain images are assigned positive
domain labels, while patch tokens extracted from retained-domain images are
assigned negative labels. Its scalar logit is converted to a domain-sensitivity
score as $s_i=\sigma(g(h_i^\ell))$, where $g(\cdot)$ denotes the patch-domain
scorer and $\sigma(\cdot)$ is the sigmoid function. The scorer is optimized
using binary cross-entropy.

\noindent\textbf{Semantic utility score.}
To avoid suppressing semantically important regions, we compute a semantic utility score for each patch token using cosine alignment with retained-class text embeddings. Specifically, for each patch token $h_i^l$, we compute
$
u_i=\max_c \cos(h_i^l,t_c),
$
where $t_c$ denotes the CLIP text embedding of retain class $c$. Patches with high text alignment are treated as semantically useful and are therefore penalized less during suppression.

\noindent\textbf{Localized suppression.}
For each patch token, we compute the localized domain effect score
$
r_i=\frac{s_i}{u_i+\epsilon},
$
where $\epsilon$ is a small numerical constant. This score increases when a patch is highly predictive of the forget domain but weakly aligned with retained semantic classes. Patches with $r_i>\tau$ are attenuated as
$
\widetilde{h}_i^l=(1-\alpha \bar r_i)h_i^l,
$
where $\alpha$ controls suppression strength and $\bar r_i$ denotes the normalized localized domain effect score. Patches below the threshold are left unchanged.

\noindent\textbf{Training and evaluation protocol.}
The CLIP backbone remains frozen throughout our experiments. Only the lightweight patch-domain scorer is trained. After training the scorer, localized suppression is applied at inference time without updating the CLIP encoder or text encoder. We evaluate performance using retained-domain accuracy, forget-domain error, and their harmonic mean, following the approximate domain unlearning protocol.

\noindent\textbf{Hyperparameters.}
Table~\ref{tab:hyperparameters} summarizes the default implementation hyperparameters used across all experiments. Unless otherwise stated, we use the same CLIP backbone, input resolution, prompt templates, and evaluation protocol across all compared methods. The attenuation strength $\alpha$, threshold $\tau$, selected intervention layer, and top-$K$ patch ratio are selected based on validation performance. We additionally employ localized top-$K$ suppression, where only the highest-scoring domain-sensitive patches are attenuated instead of uniformly modifying all patch tokens.

\begin{table}[t]
\centering
\scriptsize
\setlength{\tabcolsep}{7pt}
\renewcommand{\arraystretch}{1.05}
\caption{\textbf{Implementation Hyperparameters.} Default hyperparameter configuration used across all experiments unless otherwise specified.}
\label{tab:hyperparameters}
\begin{tabular}{lc}
\hline
\textbf{Hyperparameter} & \textbf{Value} \\
\hline

Backbone Model & CLIP ViT-B/16 \\

Input Resolution & $224 \times 224$ \\

Patch Grid Size & $14 \times 14$ \\

Number of Patch Tokens & 196 \\

Patch Feature Dimension & 768 \\

Intervention Layer & Intermediate ViT layer \\

Patch-Domain Scorer & MLP + Sigmoid \\

Domain Scorer Loss & Binary Cross-Entropy \\

Semantic Utility Metric & Cosine Similarity \\

Text Prompt Template & ``a photo of a \{class\}.'' \\

Optimizer & AdamW \\

Learning Rate & 0.0025 \\

Training Epochs & 50 \\

LDE Threshold $\tau$ & 0.3 \\

Attenuation Strength $\alpha$ & 1.0 \\

Margin Weight $\lambda$ & 0.5 \\

Top-$K$ Suppression Ratio & 10\% patches \\

Normalization Constant $\epsilon$ & $10^{-6}$ \\

CLIP Backbone Update & Frozen \\

Text Encoder Update & Frozen \\



\hline
\end{tabular}
\end{table}

\noindent\textbf{Implementation Environment.}
All experiments are implemented in PyTorch and conducted on NVIDIA H200 GPUs. The same implementation framework, CLIP backbone, input resolution, prompt templates, and evaluation protocol are used consistently across all compared methods.

\bibliography{egbib}